\documentclass[letterpaper, 10 pt, conference]{ieeeconf}  

\IEEEoverridecommandlockouts                              

\usepackage{graphics} 
\usepackage{amsmath} 
\usepackage{amssymb}  
\usepackage{xcolor}
\usepackage{float}
\usepackage{graphicx}
\usepackage{subcaption}
\usepackage[style=ieee]{biblatex}
\usepackage{bm}
\usepackage{fancyhdr}

\title{\LARGE \bf Optimizing Start Locations in Ergodic Search for Disaster Response}

\author{Ananya Rao$^{1}$, Alyssa Hargis$^{2}$, David Wettergreen$^{1}$ and Howie Choset$^{1}$
\thanks{*This material is based upon work supported by the AI Research Institutes Program funded by the National Science Foundation under AI Institute for Societal Decision Making (AI-SDM), Award No. 2229881}
\thanks{$^{1}$A. Rao, D. Wettergreen and H. Choset are with the Robotics Institute, Carnegie Mellon University, Pittsburgh, PA, USA ananyara@andrew.cmu.edu}%
\thanks{$^{2}$A. Hargis is with the United States Air Force Academy Electrical and Computer Engineering Dept.}%
}

\begin{document}

\maketitle

\begin{abstract}
In disaster response scenarios, deploying robotic teams effectively is crucial for improving situational awareness and enhancing search and rescue operations. 
The use of robots in search and rescue has been studied but the question of where to start robot deployments has not been addressed.
This work addresses the problem of optimally selecting starting locations for robots with heterogeneous capabilities—those equipped with different sensing and motion modalities—by formulating a joint optimization problem. 
To determine start locations, this work adds a constraint to the ergodic optimization framework whose minimum assigns robots to start locations. 
This becomes a little more challenging when the robots are heterogeneous - equipped with different sensing and motion modalities - because not all robots start at the same location, and a more complex adaptation of the aforementioned constraint is applied.
Our method assumes access to potential starting locations, which can be obtained from expert knowledge or aerial imagery.
We experimentally evaluate the efficacy of our joint optimization approach by comparing it to baseline methods that use fixed starting locations for all robots. 
Our experimental results show significant gains in coverage performance, with average improvements of $35.98\%$ on synthetic data and $31.91\%$ on real-world data for homogeneous and heterogeneous teams, in terms of the ergodic metric. 

\end{abstract}

\section{INTRODUCTION}

\begin{figure}[t]
    \centering
    \includegraphics[width= \linewidth]{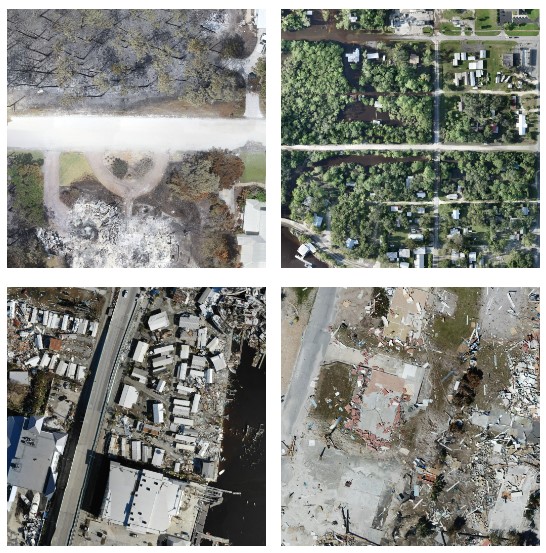}
    \caption{Aerial imagery of disaster sites listed from top left (clockwise): Musset Bayou Fire; Hurricane Idalia; Hurricane Ian; Hurricane Michael. Imagery from Center for Robot Assisted Search and Rescue~\cite{manzini2024crasarudroidslargescalebenchmark}.}
    \label{fig:disasters}
\end{figure}

Robot teams can be beneficial in disaster response scenarios, and have been used for improving situational awareness~\cite{situation_awareness}, searching collapsed buildings~\cite{disaster_robotics} and other support uses for human rescue workers~\cite{murphy2021adoption}.
Robots with heterogeneous capabilities, specifically different sensing and motion modalities, can be used to create more effective disaster response teams that can access more regions with their diverse capabilities. 
Current deployments of robots to disaster sites involve humans choosing a launch or start location for the robot, typically using prior knowledge of the region or by parsing complex aerial imagery (akin to Fig~\ref{fig:disasters}).

While a robotic team with diverse skills can be useful for disaster response or search and rescue (SAR), the efficacy of a robotic team depends on how well the robots are coordinated.
Prior work has looked at coordinating and planning paths for heterogeneous robots gathering information in or searching a disaster site~\cite{sartoretti2022spectral}.
By allocating robots to different subparts of the overall search problem, the diverse capabilities of a heterogeneous robot team can be leveraged to improve information gathering or coverage of the region. 
However, these works typically assume a single starting location for all robots, or nominally set starting locations for the robots. 
This ignores the different requirements of starting or launch locations for different types of robots.
For example, an uncrewed aerial vehicle (UAV) may need a clear, unobstructed area to take off and gain altitude, while a ground vehicle may require a stable, accessible surface to begin its operation. 
Such distinctions are crucial as they directly impact the robot's ability to initiate its tasks effectively and efficiently.
Further, the quality of search or information gathering paths possible could also depend on the starting location of a robot. 

We can automate the process of selecting suitable start locations for robots in a disaster response multi-agent team. 
In this work, we accomplish the goal of selecting starting locations that allow for effective paths for a team of robots by jointly optimizing for both start locations and robot trajectories. 


We build on our prior work in heterogeneous multi-agent ergodic search to incorporate optimizing start locations for robots. 
Ergodic search is useful for disaster response because it drives robots to spend time in areas in proportion to the expected amount of information in that region, thereby balancing exploration and exploitation~\cite{mathew2011metrics}.
This means that robots will search areas with higher likelihood of information or higher priority first, but will still explore the whole search space which accounts for the the possibility of having uncertain or inaccurate information priors. 
Spectral-based distributed heterogeneous multi-agent ergodic search, where different robots are allocated to different search subtasks using spectral analysis of the information prior driving the search, has been experimentally shown to lead to better overall coverage performance for a robotic team~\cite{sartoretti2022spectral}.
Here, we extend spectral-based distributed multi-agent ergodic search to additionally optimize for start locations for the different robots. 
Our approach assumes that we have access to a set of potential start locations to choose from, which can be obtained from expert knowledge or analysis of aerial imagery~\cite{hargis2024boo}, and focuses on selecting suitable locations from the given set. 

The efficacy of jointly optimizing start locations and robot trajectories is experimentally evaluated and compared to baseline methods of having single or multiple randomly sampled starting locations for all robots. 
We use coverage performance as our comparison metric, measured by the ergodic metric. 
We evaluate the joint optimization approach for homogeneous and heterogeneous multi-agent teams on synthetic and real-world datasets. 
Evaluated on synthetic data, our approach leads to a $39.35\%$ and $34.33\%$ improvement in coverage performance over the single random starting point baseline, for homogeneous and heterogeneous teams respectively.
Evaluated on real-world data, our approach leads to a $32.6\%$ and $29.5\%$ improvement in coverage performance over the single random starting point baseline, for homogeneous and heterogeneous teams respectively.

\section{BACKGROUND AND RELATED WORK}
\label{prior-work}

\subsection{Robotics for Disaster Response}
In recent years, the integration of robotics in disaster response and search and rescue operations has gained significant traction. 
These technologies offer valuable support to human first responders in dangerous situations, enhancing the efficiency and safety of disaster management operations~\cite{situation_awareness, murphy2021adoption}. 
Unmanned aerial vehicles (UAVs) or drones have become indispensable tools in disaster scenarios, providing aerial surveillance and data collection. 
For instance, during Hurricane Harvey in 2017, drone operators conducted several flights to assess damage and support rescue efforts~\cite{harvey}. 
However, the effectiveness of these devices heavily relies on skilled human pilots who can navigate complex environments and interpret the gathered information. 
Ground-based robots, such as the Colossus firefighting robot used in the 2019 Notre-Dame Cathedral fire, demonstrate the potential of robotic assistance in hazardous situations~\cite{wang2024robotic}. 
These machines, while sophisticated, are not fully autonomous actors but rather tools controlled by trained personnel. 

The selection of starting locations for these robots is a critical aspect that depends on human expertise. 
Factors such as the nature of the disaster, terrain characteristics, and available intelligence about potential survivors or hazards are considered when determining deployment positions~\cite{federal_emergency_mananagement_agency_fema_2021}. 
Human operators also assess the robots' capabilities and limitations, ensuring that each unit is positioned where it can be most effective given its specific design and functionality. 
Furthermore, the selection process often involves real-time coordination with other rescue teams and consideration of evolving situational dynamics, requiring continuous human judgment to adapt deployment strategies as new information becomes available.

While robotic innovations enhance the information available to rescue teams, the interpretation of data and subsequent action plans still rely heavily on human analysis and decision-making. 
Despite the increasing autonomy of disaster response robots, human operators remain crucial for their effective deployment and operation. 
From selecting initial deployment locations to guiding robots through treacherous environments and interpreting collected data, human skills and judgment continue to be at the core of successful disaster response strategies.

\subsection{Multi-Agent Informative Path Planning}
Current active search and exploration methods generally fall into one of three main categories: geometric, gradient-based, and trajectory optimization-based approaches.
Geometric methods, e.g., lawnmower patterns, can be good search strategies in order to uniformly cover a domain in which there is near-uniform probability of finding a target~\cite{choset2001coverage,Ablavsky2000}.
Since these approaches exhaustively cover the search domain, they are also the logical choice in cases where there is no \textit{a priori} information about the targets' locations.

Current informative path planning methods can generally be divided into two broad categories: gradient-based, and trajectory optimization-based approaches. 
When prior information is available, often represented as an information map or probability distribution of target locations, more sophisticated search processes can be developed. 
These methods leverage the information map to optimize the search according to specific metrics, such as minimizing the time to find all targets.

Gradient-based or "information surfing" methods guide agents towards areas of maximum information gain by following the derivative of the information map~\cite{lanillos2014multi,baxter2007multi,wong2005multi}. 
This approach can be implemented in a decentralized manner and can use potential fields to distribute agents. 
However, gradient-based methods often neglect uncertainty in the information distribution, potentially leaving areas unexplored. 
They are also sensitive to noise in the information map and tend to over-exploit local information maxima due to their greedy nature.

Optimization-based approaches treat search as an information gathering maximization problem, solved by planning paths for agents. 
Recent coverage methods use sampling-based path planning, selecting the best path based on a cost metric \cite{ayvaliRAL2016,ayvali2017ergodic,miller2015,mathew2011metrics}. 
These approaches can incorporate both predicted information distribution and uncertainty into the cost function. 
However, they often don't scale well for large multi-agent systems due to their centralized nature. 
While the number of paths to sample grows exponentially with agent count, increasing samples linearly with team size has shown good results experimentally~\cite{ayvaliRAL2016,ayvali2017ergodic}.

\subsection{Ergodic Trajectory Optimization}
Ergodic search processes \cite{mathew2011metrics} produce trajectories that drive agents to spend time in areas of the domain in proportion to the expected amount of information present in those areas. 
The spatial time-average statistics of an agent's trajectory (trajectory is represented as \mbox{$\gamma_i:(0,t]\rightarrow \mathcal{X}$}), specifies the amount of time spent at position
$\bm{x}\in \mathcal{X}$, where $\mathcal{X}\subset {\rm I\!R^d}$ is the $d$-dimensional search domain.
For $N$ agents, the joint spatial time-average statistics of the set of agents trajectories $\{\gamma_i\}_{i=1}^N$ is defined as~\cite{mathew2011metrics}

\begin{equation}
C^t(\bm{x},\gamma(t))=\frac{1}{Nt}\sum_{i=1}^{N}\int_{0}^{t} \delta(\bm{x}-\gamma_i(\tau)) \,\, d\tau,
\end{equation}

\noindent where $\delta$ is the Dirac delta function. 

The agents' trajectories are optimized by matching the spectral decompositions of the time-averaged trajectory statistics and the information distribution over the search domain.
This is accomplished by minimizing the ergodic metric $\Phi (\cdot)$, which is the weighted sum of the difference between the spectral coefficients of these two distributions~\cite{mathew2011metrics}:

\begin{equation}
\Phi(\gamma(t))=\sum_{k=0}^{m} \alpha_k \left| c_k (\gamma(t)) - \xi_k \right|^2,
\label{IROS2021-eq:ergodicMetric}
\end{equation}

\noindent where $c_k$ and $\xi_k$ are the Fourier coefficients of the time-average statistics of the set of agents' trajectories $\gamma (t)$ and the desired spatial distribution of agents respectively, $\alpha_k$ are the weights of each coefficient difference, and $m$ is the number of Fourier coefficients being considered. 

The goal of ergodic coverage is to generate optimal controls $\bm{u}^*(t)$ for each agent, whose dynamics is described by a function $f\colon \mathcal{Q} \times \mathcal{U} \to \mathcal{TQ} $, such that

\begin{equation}
\begin{array}{rcl}
\label{IROS2021-eq:robot_eqn}
\bm{u}^*(t) & = & \operatorname*{arg\,min}_{\bm{u}} \Phi(\gamma(t)),\\[0.3cm]
\mbox{subject to } \dot{\bm{q}} & = & f(\bm{q}(t),\bm{u}(t)),\\[0.15cm]
\left\|\bm{u}(t)\right\| & \leq & u_{max}
\end{array}
\end{equation}

\noindent where $\bm{q}\in \mathcal{Q}$ is the state and $\bm{u}\in \mathcal{U}$ denotes the set of controls.

Prior work extended ergodic search to present an automated approach to creating search subtasks given an information map to explore~\cite{sartoretti2022spectral}. 
Specifically, this work used Fourier decomposition to distribute different spectral scales of the search problem to different agents, based on agent sensing and motion model.
This paper uses spectral-based distributed heterogeneous multi-agent search as the path planner. 

\section{JOINTLY OPTIMIZING START LOCATIONS AND ROBOT TRAJECTORIES}
\label{approach}
We approach the problem of selecting suitable starting locations for different robots in a multi-agent team by formulating a joint optimization problem to solve for the robots' starting locations and their search paths. 
We extend the ergodic trajectory optimization formulation to incorporate starting locations for each robot as additional optimization variables.
Mathematically, this looks like adding an additional constraint to the ergodic trajectory optimization formulation in Eqn~\ref{IROS2021-eq:robot_eqn}, 

\begin{equation}
    \{\gamma_i\}_{i=1}^N(0) \in \mathcal{X}_0 
    \label{eqn:starting_constraint}
\end{equation}

where $\mathcal{X}_0$ is the set of all viable or potential starting locations for the agent. 
Our approach assumes knowledge of possible or viable starting locations to choose from, that is, $\mathcal{X}_0$ is available to the algorithm.
In practice, this set of viable starting locations can come from processing satellite or aerial imagery or from expert knowledge.

We consider two varieties of multi-agent teams: homogeneous, where all agents have the same motion and sensing models, and heterogeneous, where agents have different motion and sensing models. 
In the case of a homogeneous multi-agent team, we assume that all of the robots share the same set of possible starting locations, as they have the same capabilities. 
Therefore, Eqn~\ref{eqn:starting_constraint} can be directly added as a constraint to ergodic trajectory optimization (Eqn~\ref{IROS2021-eq:robot_eqn}). 
For a heterogeneous multi-agent team, different agents can have different sets of possible starting locations.
Specifically, if there are $M$ types of agents in the team, there are $M$ sets of viable starting locations, $\{X_0^i\}_{i=1}^M$, and the constraint to be added to the ergodic trajectory optimization problem is,

\begin{equation}
\begin{array}{lcl}
    \gamma_i(0) &\in X_0^i &\forall i \in [1, n_k] \\ &&\forall k \in [1, M]
    \label{eqn:starting_constraint_het}
\end{array}
\end{equation}

where $n_k$ is the number of agents of type $k$ and $\sum_{k=1}^M n_k = N$.

\section{RESULTS AND DISCUSSION}
\label{results}

\subsection{Experiment Details}
\begin{figure}[t]
    \centering
    \begin{subfigure}[b]{0.15\textwidth}
         \centering
         \includegraphics[width=\textwidth]{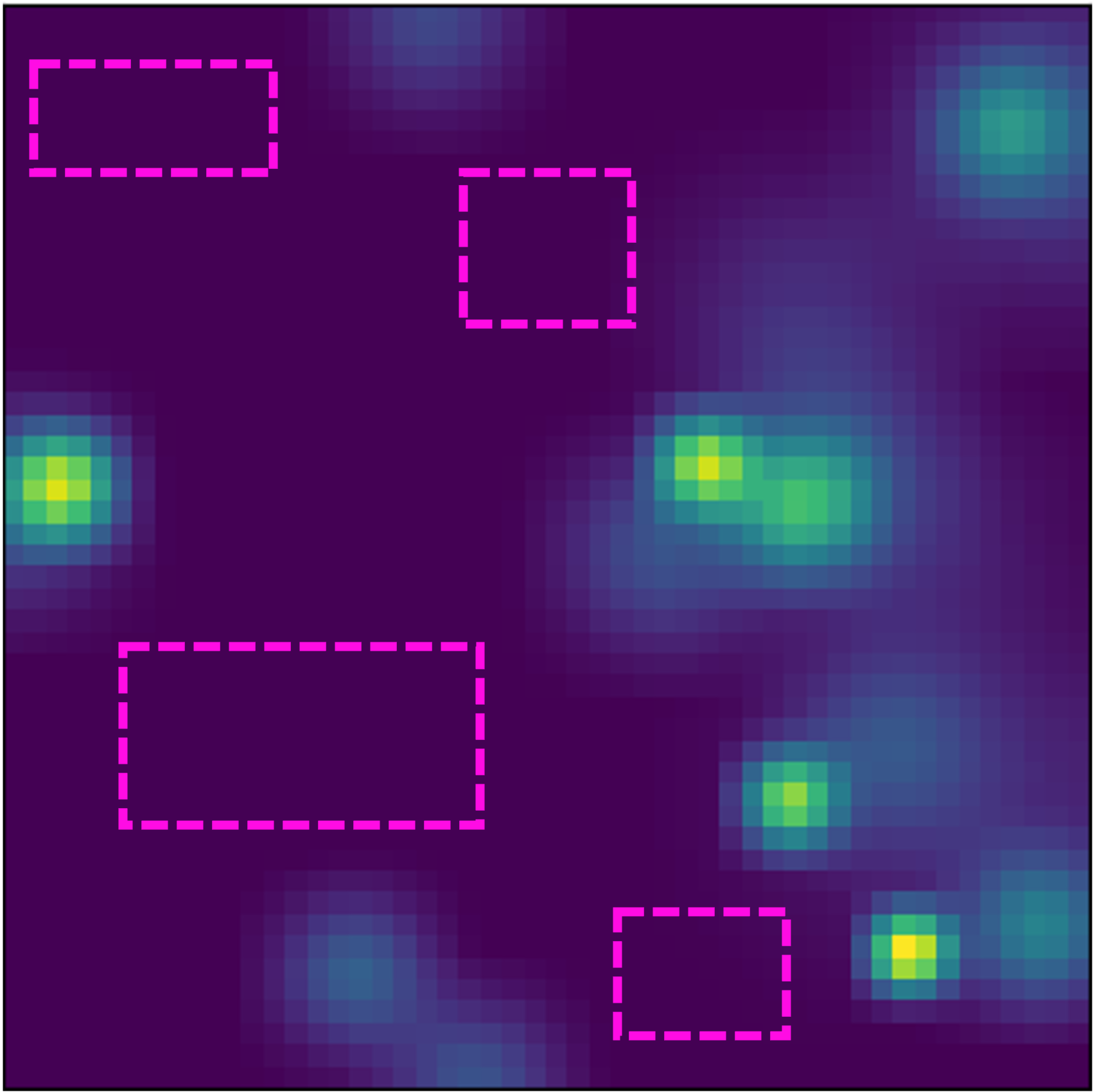}
     \end{subfigure}
     \begin{subfigure}[b]{0.15\textwidth}
         \centering
         \includegraphics[width=\textwidth]{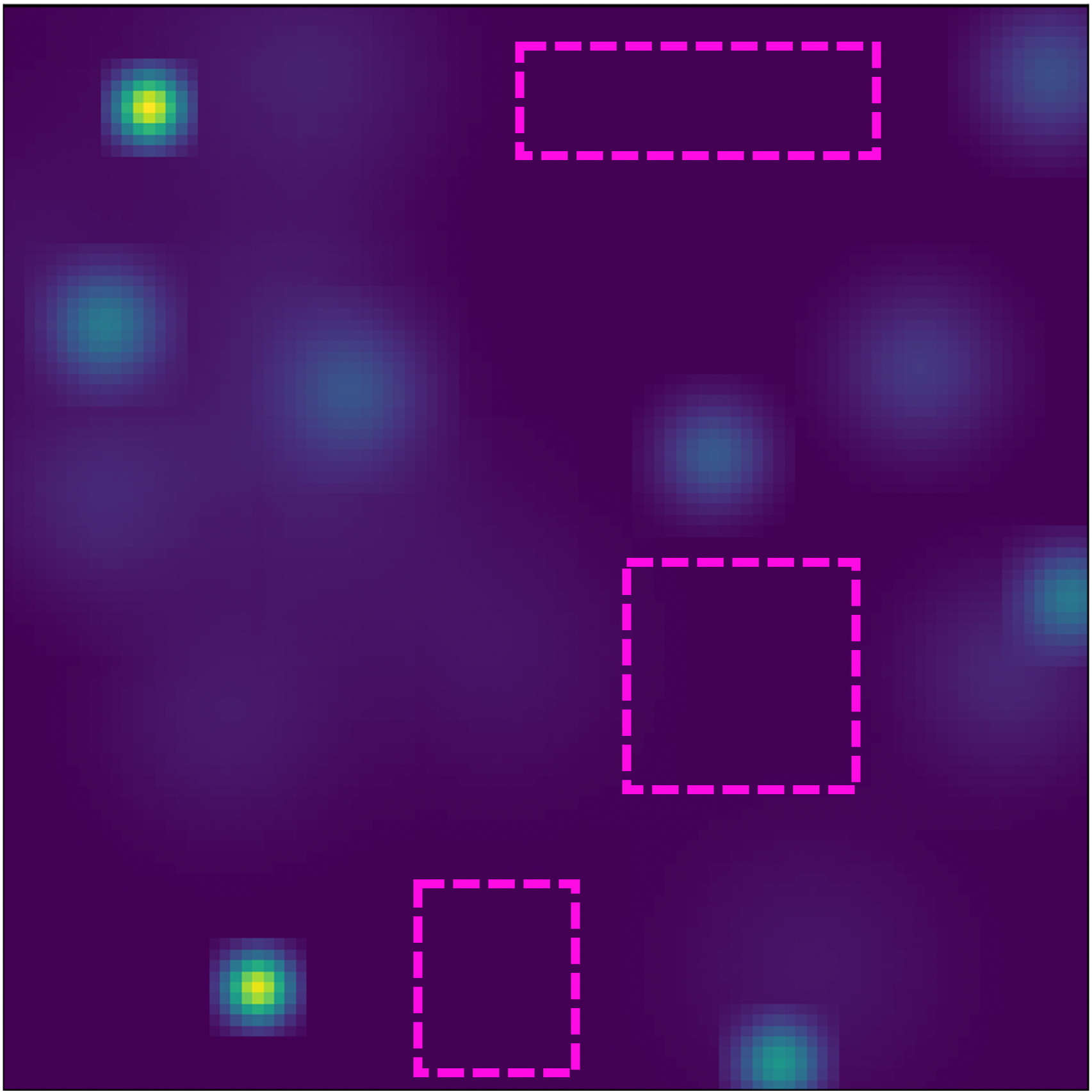}
     \end{subfigure} 
     \begin{subfigure}[b]{0.15\textwidth}
         \centering
         \includegraphics[width=\textwidth]{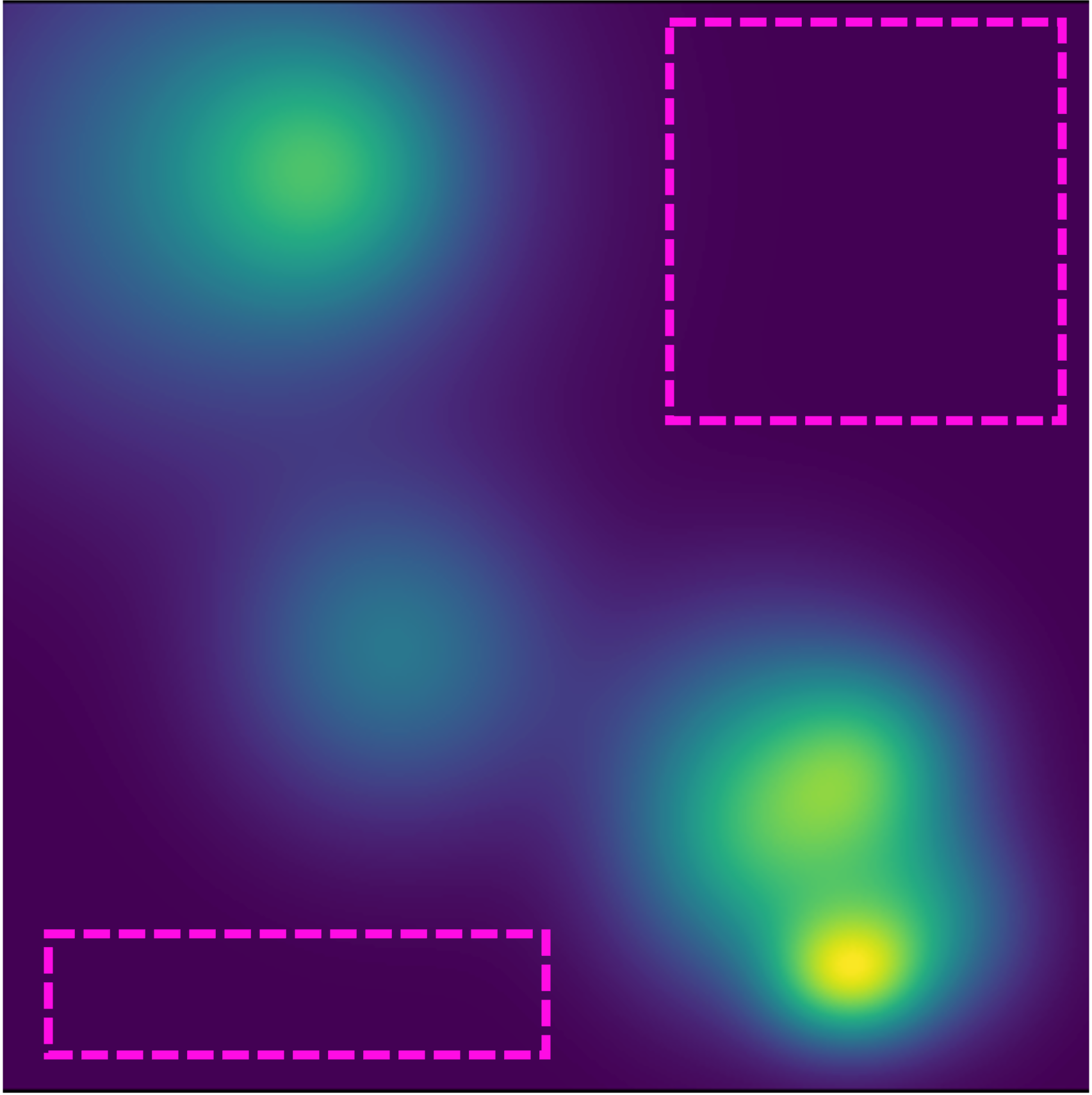}
     \end{subfigure}
 \caption{Synthetic dataset examples: Yellow indicates high information, purple denotes low information, and pink dashed lines highlight areas of potential starting locations.} 
    \label{fig:synthetic_data}
\end{figure}

\paragraph{Datasets}
Our experimental analysis uses two datasets, one synthetically generated, and one developed from analysis of aerial imagery of real-world disaster sites.
Each dataset consists of two key aspects: information maps that encode areas of interest or the expected amount of information at each point in the search region, and areas of potential starting locations. 
For the synthetically generated dataset, information maps were generated using Gaussian mixture models and areas of potential starting locations were arbitrarily chosen. 
Example maps from this dataset are shown in Fig~\ref{fig:synthetic_data}.

\begin{figure}[t]
    \centering
    \includegraphics[width=\linewidth]{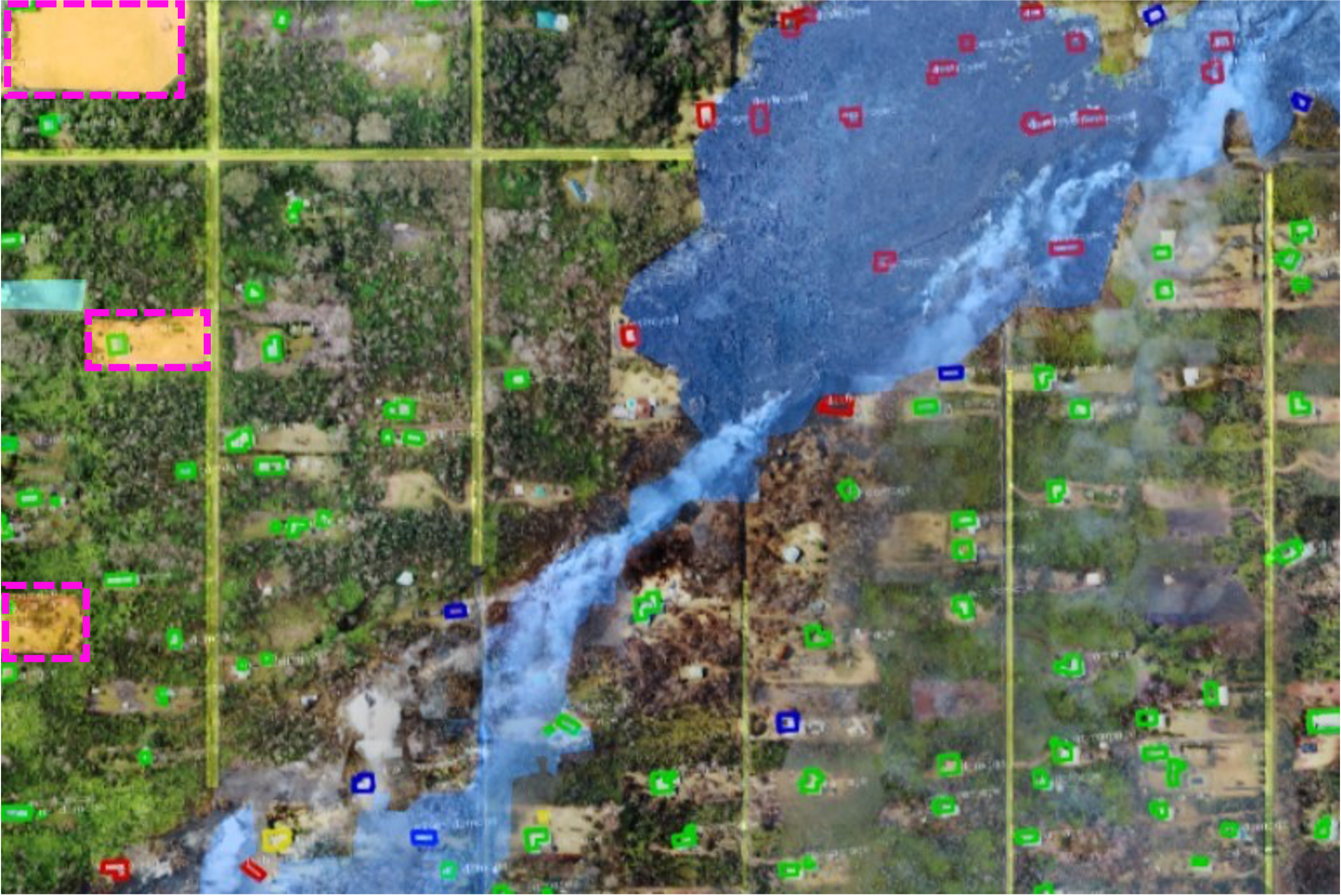} 
    \includegraphics[width=\linewidth]{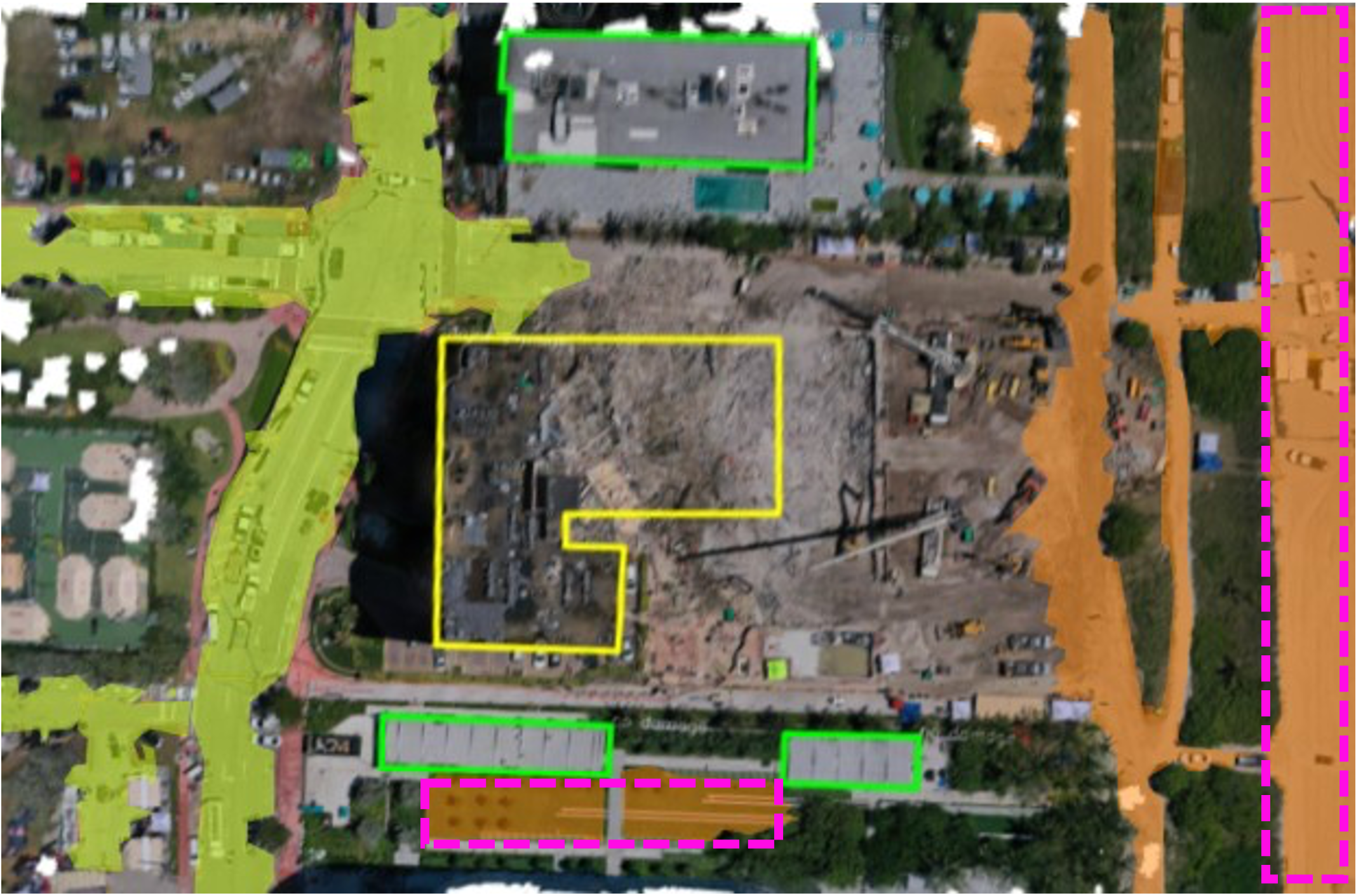}
    \caption{Real-world dataset examples: Building damage is shown on aerial imagery with red for severe, yellow for moderate, and green for minor damage; pink dashed lines mark areas of potential starting locations.}
    \label{fig:realworld_data}
\end{figure}

For the real-world dataset, information maps are generated using building damage assessment overlays from the CRASAR dataset~\cite{manzini2024crasarudroidslargescalebenchmark}, with higher expected information over more damaged buildings, and lower expected information in areas without buildings. 
Areas of potential starting locations were determined by analysis of the aerial imagery~\cite{hargis2024boo}.
Example maps showing building damage assessment overlays and areas of potential starting location are shown in Fig~\ref{fig:realworld_data}.

\paragraph{Agents}
Each agent's sensor is modeled as a Gaussian distribution centered on the agent's location, with detection likelihood defined by the Gaussian probability density function at each point within the sensor's footprint. 
We consider two types of sensors: a low-range, high-fidelity sensor with a narrow Gaussian spread and high detection probability, and a high-range, low-fidelity sensor with a wider Gaussian spread and lower detection probability. 
In addition to varying sensor models, we also account for two motion models: omnidirectional agents, such as quad-rotor UAVs, which are modeled as simple first-order integrators, and differential drive agents, like wheeled ground vehicles, which are modeled with curved paths constrained by a maximum curvature.

\paragraph{Baseline Approaches}
We compare jointly optimizing starting locations and robot trajectories to three different baseline methods.
First, we consider a planner where robot trajectories are optimized given a fixed single random starting location for all robots. 
In the second baseline method, robot trajectories are optimized given multiple randomly selected fixed starting locations, one for each robot in the multi-agent exploration team.
Finally, we consider a method where a single starting location is jointly optimized with robot trajectories.
In this case, all of the robots use the same optimized starting location. 
Spectral-based distributed ergodic search is used in all cases to plan paths for the multi-agent search teams. 
We note that while spectral-based distributed ergodic search was developed for planning paths for heterogeneous multi-agent teams, it can also be used for homogeneous teams. 
However, the benefits of the algorithm are clearer for heterogeneous teams. 

\subsection{Experimental Results}
\begin{figure}[t]
    \centering
    \includegraphics[width=\linewidth]{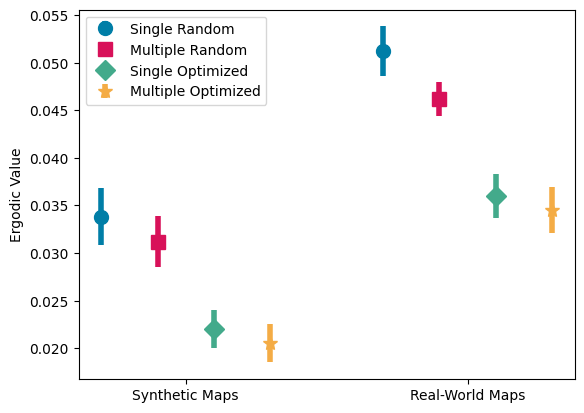}
    \caption{Homogeneous Multi-Agent Team Experiments: Evaluation of planning with a single random starting location, multiple random starting locations, single optimized starting location, and multiple optimized starting locations in terms of coverage performance using the ergodic metric (lower is better) on synthetic and real-world information maps.}
    \label{fig:res-synth}
\end{figure}

\begin{figure}[t]
    \centering
    \includegraphics[width=\linewidth]{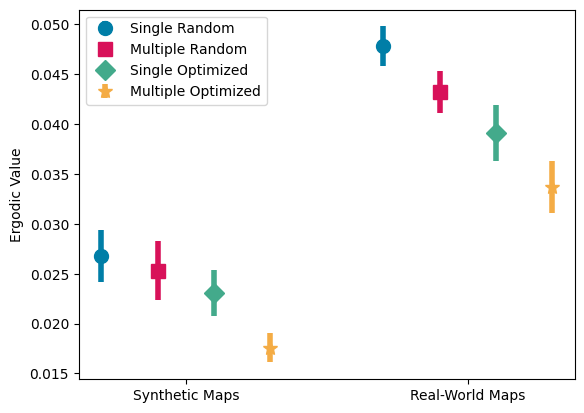}
    \caption{Heterogeneous Multi-Agent Team Experiments: Evaluation of planning with a single random starting location, multiple random starting locations, single optimized starting location, and multiple optimized starting locations in terms of coverage performance using the ergodic metric (lower is better) on synthetic and real-world information maps.}
    \label{fig:res-rw}
\end{figure}

We first look at the empirical results for experiments run with homogeneous teams, that is, a set of robots with the same sensing and motion capabilities (seen in Fig~\ref{fig:res-synth}).
Planning multi-agent search trajectories while choosing a single fixed random starting location has the worst coverage performance in terms of the ergodic metric, as the chosen starting location may or may not be well-suited for planning paths to gather information across the search region. 
Coverage performance does slightly improve when multiple starting locations are chosen at random.
This is because increasing the number of starting locations increases the likelihood of choosing a starting location that is well-placed, however, given that this is a benefit gotten by random chance, the performance increase is limited (average improvement over planning with a single random starting location is $7.69\%$ on synthetic maps and $9.77\%$ on real-world maps).

We further see that planning with a single optimized starting location improved performance by $34.91\%$ on synthetic maps and $29.69\%$ on real-world maps over planning with a single random starting point, as jointly optimizing starting location and robot trajectories results in a well-suited starting location for effective robot trajectories. 
Finally, we see that planning with multiple optimized starting locations results in similar coverage performance to planning with a single optimized starting location (i.e. $39.35\%$ improvement for synthetic maps and $32.6\%$ improvement for real-world maps over planning with a single random starting location). 
For multi-agent homogeneous exploration teams, the same starting location could serve all of the robots well, due to the robots having the same capabilities. 
This leads to the similar coverage performance of both planning with a single optimized starting location and multiple optimized starting locations. 
Empirically, we see that when there are multiple similarly "good" starting locations, having multiple optimized starting locations can lead to better overall team performance, since agents are initially spatially spread across the search space better. 
This could explain the slight improvement in performance over using a single optimized start location. 
All coverage performance statistics are averaged across 50 experimental trials each (5 runs each on 10 different maps). 

Additionally, we evaluate the same four planning methods using heterogeneous multi-agent teams, that is exploration teams consisting of robots with different sensing and motion modalities.
The results of these experiments are shown in Fig~\ref{fig:res-rw}.
For heterogeneous teams, planning with a single random starting location has the worst coverage performance in terms of the ergodic metric, as a single randomly chosen starting location cannot be well-suited for search paths for heterogeneous agents.
Planning with multiple random starting locations leads to some slight improvements in coverage performance ($5.52\%$ on synthetic maps and $9.62\%$ on real-world maps).
However, performance improvements are limited by the slim likelihood of randomly selecting a well-suited starting location for each robot, given that each type of robot may be better suited for a different type of starting location.

Further, we see that planning with a single optimized starting location does better than single or multiple random starting locations ($13.81\%$ on synthetic maps and $18.2\%$ on real-world maps).
However, given that the optimal starting location will be different for different types of robots, a single starting location may require choosing a location that is suboptimal, or in the worst case adversarial, for some of the robot types in the heterogeneous team. 
Finally, when multiple starting locations are jointly optimized with robot trajectories, a well-suited starting location is chosen for each type of robot, which leads to larger improvements in coverage performance ($34.33\%$ on synthetic maps and $29.5\%$ on real-world maps).
All coverage performance statistics are averaged across 25 experimental trials each (5 runs each on 5 different maps). 

Empirically we see the largest improvement in coverage performance over planning with a single random starting location when planning with multiple optimized starting locations. 
This improvement in coverage performance is likely due to the joint optimization method allowing for the selection of starting locations that are well-suited to the search trajectories that robots are following.
When restricted to pre-determined randomly chosen starting locations, the efficacy of a robot's trajectory is limited to what is reachable from that starting point within the available search time and the robot's exploration budget, which could be limited by battery, communication limits, or mission duration. 
By making starting location selection a part of the overall optimization problem, robots can begin their exploration from locations that are better suited to the regions of the disaster site that they will be exploring, without having to rely on human operators to pick those starting locations.

\section{CONCLUSIONS}
\label{conclusion}
Our approach to optimizing both starting locations and trajectories for robotic teams offers a significant advancement in disaster response and search and rescue operations. 
By extending the ergodic trajectory optimization framework to include starting positions as optimization variables, we enable more effective deployment of robots with heterogeneous capabilities. 
This joint optimization not only improves the initial positioning of robots based on their specific needs and requirements but also enhances their overall performance in terms of information gathering and area coverage. 
Our experimental results demonstrate that the proposed method achieves substantial improvements over traditional approaches that use fixed or single starting locations, with notable performance gains for both homogeneous and heterogeneous robot teams, on synthetic and real-world datasets.

We note that there are practical considerations to be taken into account when operating with multiple starting locations.
For example, this work assumes that all robots will start operation at the same time, even when starting from different locations.
In reality, due to travel times between start locations and other challenges this may not be possible.
Further research is required to investigate the effect of offset operation start times on overall team performance. 
Further, our approach does not incorporate the operational costs of having personnel at multiple robot deployment sites, as well as the cost of traveling between these sites. 
Future work could look in to capturing these costs within the optimization framework.

Future work could focus on several key areas to further enhance the capabilities and application of our method. 
First, incorporating real-time environmental changes and dynamic obstacles into the optimization process could improve the adaptability and robustness of robot deployments in rapidly evolving disaster scenarios. 
Additionally, exploring more sophisticated metrics for evaluating information-gathering efficiency and integrating machine learning techniques for adaptive trajectory planning could provide further improvements. 
Expanding the approach to include collaborative decision-making frameworks, where robots can adjust their starting locations and paths based on ongoing observations and interactions, represents another promising direction for future research.

\addtolength{\textheight}{-12cm}   







\printbibliography

@book{disaster_robotics,
    author = {Robin R Murphy},
    title = {Disaster Robotics},
    publisher = {MIT Press},
    year = {2014}
}

@misc{manzini2024crasarudroidslargescalebenchmark,
      title={CRASAR-U-DROIDs: A Large Scale Benchmark Dataset for Building Alignment and Damage Assessment in Georectified sUAS Imagery}, 
      author={Thomas Manzini and Priyankari Perali and Raisa Karnik and Robin Murphy},
      year={2024},
      eprint={2407.17673},
      archivePrefix={arXiv},
      primaryClass={cs.CV},
      url={https://arxiv.org/abs/2407.17673}, 
}

@article{mathew2011metrics,
  title={Metrics for ergodicity and design of ergodic dynamics for multi-agent systems},
  author={Mathew, George and Mezi{\'c}, Igor},
  journal={Physica D: Nonlinear Phenomena},
  volume={240},
  number={4},
  pages={432--442},
  year={2011},
  publisher={Elsevier}
}

@article{situation_awareness, title={Technologies Enabling Situational Awareness During Disaster Response: A Systematic Review}, volume={16}, DOI={10.1017/dmp.2020.196}, number={1}, journal={Disaster Medicine and Public Health Preparedness}, author={Kedia, Tara and Ratcliff, Jeremy and O’Connor, Megan and Oluic, Sophia and Rose, Michelle and Freeman, Jeff and Rainwater-Lovett, Kaitlin}, year={2022}, pages={341–359}}

@article{murphy2021adoption,
  title={Adoption of Robots for Disasters: Lessons from the Response to COVID-19},
  author={Murphy, Robin R and Gandudi, Vignesh BM and Adams, Justin and Clendenin, Angela and Moats, Jason and others},
  journal={Foundations and Trends{\textregistered} in Robotics},
  volume={9},
  number={2},
  pages={130--200},
  year={2021},
  publisher={Now Publishers, Inc.}
}

@InProceedings{sartoretti2022spectral,
author="Sartoretti, Guillaume
and Rao, Ananya
and Choset, Howie",
editor="Matsuno, Fumitoshi
and Azuma, Shun-ichi
and Yamamoto, Masahito",
title="Spectral-Based Distributed Ergodic Coverage for Heterogeneous Multi-agent Search",
booktitle="Distributed Autonomous Robotic Systems",
year="2022",
publisher="Springer International Publishing",
address="Cham",
pages="227--241",
isbn="978-3-030-92790-5"
}

@article{hargis2024boo,  author={Hargis, Alyssa and Rao, Ananya and Choset, Howie},  booktitle={2024 IEEE International Symposium on Safety, Security, and Rescue Robotics (SSRR)},   title={Search and Rescue Base of Operation Prioritization with Aerial Orthomosaics},   year={2024},  volume={},  number={},  pages={},  doi={}}

@incollection{Ablavsky2000,
    author = {Ablavsky, Vitaly and Snorrason, Magnus},
    title = {{Optimal search for a moving target - A geometric approach}},
    booktitle = {AIAA Guidance, Navigation, and Control Conference and Exhibit},
    publisher = {AIAA},
    year = {2000}
}

@article{choset2001coverage,
  title={Coverage for robotics--A survey of recent results},
  author={Choset, Howie},
  journal={Annals of mathematics and artificial intelligence},
  volume={31},
  number={1},
  pages={113--126},
  year={2001},
  publisher={Springer}
}

@inproceedings{ayvali2017ergodic,
  title={Ergodic coverage in constrained environments using stochastic trajectory optimization},
  author={Ayvali, Elif and Salman, Hadi and Choset, Howie},
  booktitle={International Conference on Intelligent Robots and Systems},
  pages={5204--5210},
  year={2017},
  organization={IEEE}
}

@article{ayvaliRAL2016,
  title={Utility-Guided Palpation for Locating Tissue Abnormalities},
  author={Ayvali, Elif and Ansari, Alexander and Wang, Long and Simaan, Nabil and Choset, Howie},
  journal={Robotics and Automation Letters},
  year={2017},
}

@article{miller2015,
  title={Ergodic exploration of distributed information},
  author={Miller, Lauren M and Silverman, Yonatan and MacIver, Malcolm A and Murphey, Todd D},
  journal={IEEE Transactions on Robotics},
  volume={32},
  number={1},
  pages={36--52},
  year={2016},
  publisher={IEEE}
}

@inproceedings{wong2005multi,
  title={Multi-vehicle Bayesian search for multiple lost targets},
  author={Wong, El-Mane and Bourgault, Fr{\'e}d{\'e}ric and Furukawa, Tomonari},
  booktitle={International Conference on Robotics and Automation},
  pages={3169--3174},
  year={2005},
  organization={IEEE}
}

@article{lanillos2014multi,
  title={Multi-UAV target search using decentralized gradient-based negotiation with expected observation},
  author={Lanillos, Pablo and Gan, Seng Keat and Besada-Portas, Eva and Pajares, Gonzalo and Sukkarieh, Salah},
  journal={Information Sciences},
  volume={282},
  pages={92--110},
  year={2014},
  publisher={Elsevier}
}

@incollection{baxter2007multi,
  title={Multi-robot search and rescue: A potential field based approach},
  author={Baxter, Joseph L and Burke, EK and Garibaldi, Jonathan M and Norman, Mark},
  booktitle={Autonomous robots and agents},
  pages={9--16},
  year={2007},
  publisher={Springer}
}

@INPROCEEDINGS{harvey,

  author={Fernandes, Odair and Murphy, Robin and Adams, Justin and Merrick, David},

  booktitle={2018 IEEE International Symposium on Safety, Security, and Rescue Robotics (SSRR)}, 

  title={Quantitative Data Analysis: CRASAR Small Unmanned Aerial Systems at Hurricane Harvey}, 

  year={2018},

  volume={},

  number={},

  pages={1-6},

  doi={10.1109/SSRR.2018.8468647}}

@article{wang2024robotic,
  title={Robotic Firefighting: A Review and Future Perspective},
  author={Wang, Meng and Chen, Xinghao and Huang, Xinyan},
  journal={Intelligent Building Fire Safety and Smart Firefighting},
  pages={475--499},
  year={2024},
  publisher={Springer}
}

@report{federal_emergency_mananagement_agency_fema_2021,
	title = {{FEMA} Preliminary Damage Assessment Guide},
	institution = {{US} Department of Homeland Security},
	author = {Federal Emergency Mananagement Agency},
	date = {2021-08},
	langid = {english},
}

\end{document}